\pdfoutput=1

\documentclass[11pt]{article}

\usepackage[]{acl}

\usepackage{times}
\usepackage{latexsym}

\usepackage[T1]{fontenc}

\usepackage[utf8]{inputenc}

\usepackage{microtype}

\usepackage{graphicx}
\usepackage{url}
\usepackage[utf8]{inputenc}
\inputencoding{utf8}
\usepackage{CJKutf8}
\usepackage{multirow}
\usepackage{lipsum}
\usepackage{url}
\usepackage{amsmath}
\usepackage{bm}
\usepackage{todonotes}
\usepackage{wrapfig}
\usepackage{graphicx}
\usepackage{float}
\usepackage{tikz}
\usetikzlibrary{automata,positioning,arrows}
\usetikzlibrary{decorations.text}
\usepackage{bbm}
\usepackage{threeparttable}

\usepackage{caption}
\usepackage{latexsym}
\usepackage{amssymb}
\usepackage{amsmath}
\usepackage{subcaption}
\usepackage{algorithm,algpseudocode}
\usepackage{footnote}
\makesavenoteenv{tabular}
\usepackage[export]{adjustbox}
\usepackage{amsbsy}

\title{ITEm: Unsupervised Image-Text Embedding Learning for eCommerce}

\author{
  Baohao Liao\thanks{\:\:Work done while as a researcher in eBay.} \:\:\: Michael Kozielski \:\:\: Sanjika Hewavitharana\\ 
  \textbf{Jiangbo Yuan} \:\:\:\:\:\:\:\: \textbf{Shahram Khadivi} \:\:\:\:\:\:\:\: \textbf{Tomer Lancewicki} \\
  eBay Inc. \\
  \texttt{\href{mailto:liaobaohao@gmail.com}{liaobaohao@gmail.com}}
}

\begin{document}
\maketitle

\begin{abstract}
Product embedding serves as a cornerstone for a wide range of applications in eCommerce. The product embedding learned from multiple modalities shows significant improvement over that from a single modality, since different modalities provide complementary information. However, some modalities are more informatively dominant than others. How to teach a model to learn embedding from different modalities without neglecting information from the less dominant modality is challenging. We present an \textbf{i}mage-\textbf{t}ext \textbf{em}bedding model (\textbf{ITEm}), an unsupervised learning method that is designed to better attend to image and text modalities. We extend BERT by (1) learning an embedding from text and image without knowing the regions of interest; (2) training a global representation to predict masked words and to construct masked image patches without their individual representations. We evaluate the pre-trained ITEm on two tasks: the search for extremely similar products and the prediction of product categories, showing substantial gains compared to strong baseline models.
\end{abstract}

\section{Introduction}
The global eCommerce market has grown into an unprecedented scale, expected to total \$4.89 trillion in 2021 and taking up 21.8\% of total retail sales \cite{michkee}. For various eCommerce applications, like search and classification, product embedding plays an important role. Some online shopping sites employ visual similarity to improve recall and relevance of search and recommendation results \cite{periyathambi2023systems, jing2015visual, shankar2017deep, yang2017visual}. The visual similarity is useful in some domains, like fashion and furniture. Some online shopping sites prefer to learn embedding from product title \cite{bianchi2021query2prod2vec, 10.1145/2783258.2788627}.

\begin{figure}
  \centering
  \includegraphics[width=0.48\textwidth]{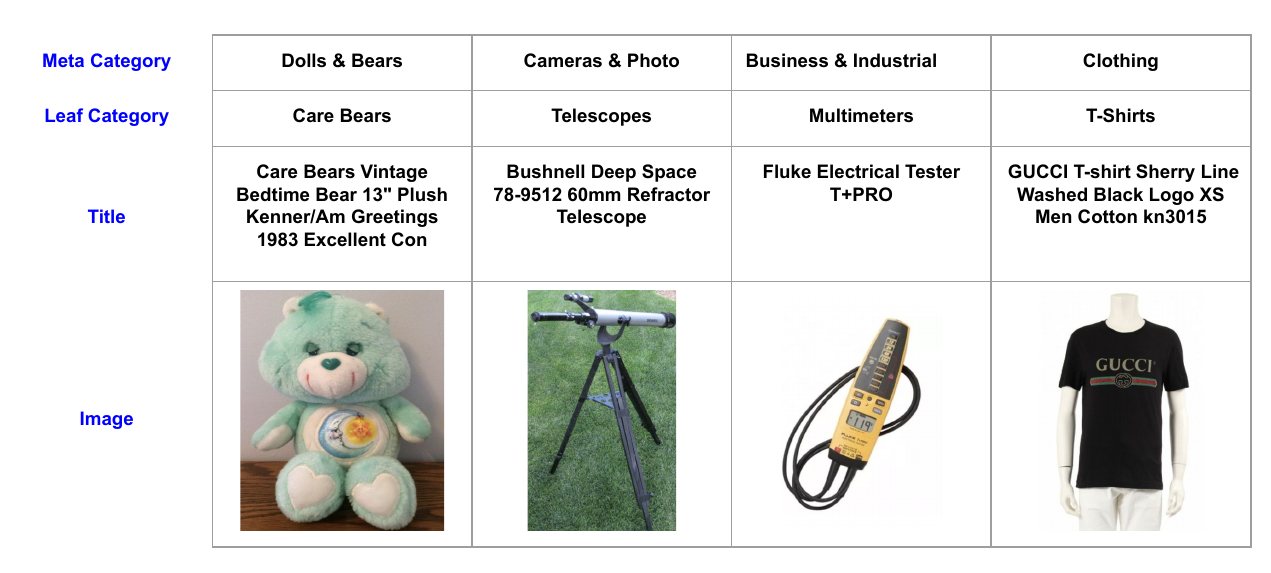}
  \caption{Examples from ITOP.}
  \label{fig:data_examples}
\end{figure}

Figure \ref{fig:data_examples} shows some product examples from our own collected dataset, \textbf{i}mage-\textbf{t}ext \textbf{o}nline \textbf{p}roduct dataset (\textbf{ITOP}). It is quite obvious that both product image and title contribute to distinguishing these products. Since both product image and title offer useful information, sometimes complementary information, it is necessary to learn a global embedding from these two modalities rather than only from a single modality.

It is challenging to learn a global embedding from different modalities. From our observation, titles contain more dominant information than images for most products. Like some examples in Figure \ref{fig:data_examples}, their titles can even contain their leaf categories. If we don't carefully design the embedding learning strategy and only casually fuse the embeddings from different modalities, the global embedding is very likely to be dominated by a single modality. Even though this kind of global embedding works for most of our products, it fails for those products whose images are more important or equally important as titles. 

In this work, we propose a new model to learn image-text embedding, i.e. ITEm, to learn a fine-grained embedding from both product image and title in an unsupervised way. Our contributions are summarized as follows:

\begin{itemize}
    \item We propose a method to learn fine-grained representation from product images and titles for eCommerce, trying to ease the embedding over-dominance problem caused by a single modality that contains dominant information.
    \item To evaluate the fine-grained embedding, we collect a large-scale image-text product dataset and annotate it for the search for extremely similar products. We evaluate our model, ITEm, on this difficult analogy task and classification task against other SOTA uni-modal and multi-modal models.
    \item We learn the image-title embedding without knowing the regions of interest, which broadens the application possibility of our method to other domain datasets.
\end{itemize}

\section{Related Work}
Image and text are two common modalities that can easily obtained from online shopping sites. Bridging them in an efficient way has a long history. Various tasks, such as image captioning \cite{you2016image}, textual grounding \cite{kazemzadeh2014referitgame, plummer2015flickr30k}, visual question answering \cite{antol2015vqa, goyal2017making} and visual reasoning \cite{suhr2018corpus, zellers2019recognition} have been proposed. To solve these tasks, various models have been developed. These models can be split into two types: a single-stream model \cite{li2019visualbert, su2019vl, qi2020imagebert} that fuses vision-and-text information at the very beginning and a two-stream model \cite{lu2019vilbert, tan2019lxmert} that consists of an image encoder, a text encoder and a multi-modal fusion module at the end.

Our ITEm is a single-stream model with only one encoder for both vision and text. ITEm is more efficient than the two-stream model, because we don't need to get the embedding for different modalities in a sequential manner. There are two main differences between ITEm and other multi-modal models: 
\begin{itemize}
    \item Inspired by \citet{dosovitskiy2020image}, ITEm uses the whole image as input without the detection of the regions of interest, while the other methods \cite{li2019visualbert, su2019vl, qi2020imagebert, lu2019vilbert, tan2019lxmert} employ a pre-trained model to detect the regions of interest and apply these regions as input. The detection of regions of interest is beneficial for some specific tasks, like image captioning, visual question answering and so on. However, if the dataset of the downstream task is out-of-domain, the performance of the object detection in this dataset is poor, which restricts the application of this kind of model on poorly annotated tasks. 
    \item We collect a new dataset to evaluate the pre-trained model. Some previous works \cite{li2019visualbert, su2019vl, qi2020imagebert, lu2019vilbert, tan2019lxmert} evaluate the pre-trained models by fine-tuning them on downstream tasks. There is a mismatch between pre-training and fine-tuning, making it difficult to evaluate the pre-training method. We evaluate the pre-trained ITEm directly on a fine-grained product recommendation task.
\end{itemize}

\section{Image-Text Online Product Data}
How to evaluate the pre-trained ITEm is a key point in our paper. Most cross-modal pre-trained models \cite{lu2019vilbert, li2019visualbert, su2019vl, tan2019lxmert, desai2020virtex} use the Conceptual Captions dataset \cite{sharma2018conceptual} or the Microsoft COCO Captions dataset \cite{chen2015microsoft} as pre-training datasets. These models are then fine-tuned on retrieval tasks or question-answering tasks for evaluation. There is a mismatch between pre-training and fine-tuning methods, which makes it difficult to demonstrate the effectiveness of a pre-training method. Even though some works, like \citet{qi2020imagebert}, evaluate the embedding from a pre-trained model on zero-shot retrieval tasks, there is still a lack of dataset to evaluate the fine-grained embedding. 

\begin{figure}[t]
  \centering
  \includegraphics[width=0.45\textwidth]{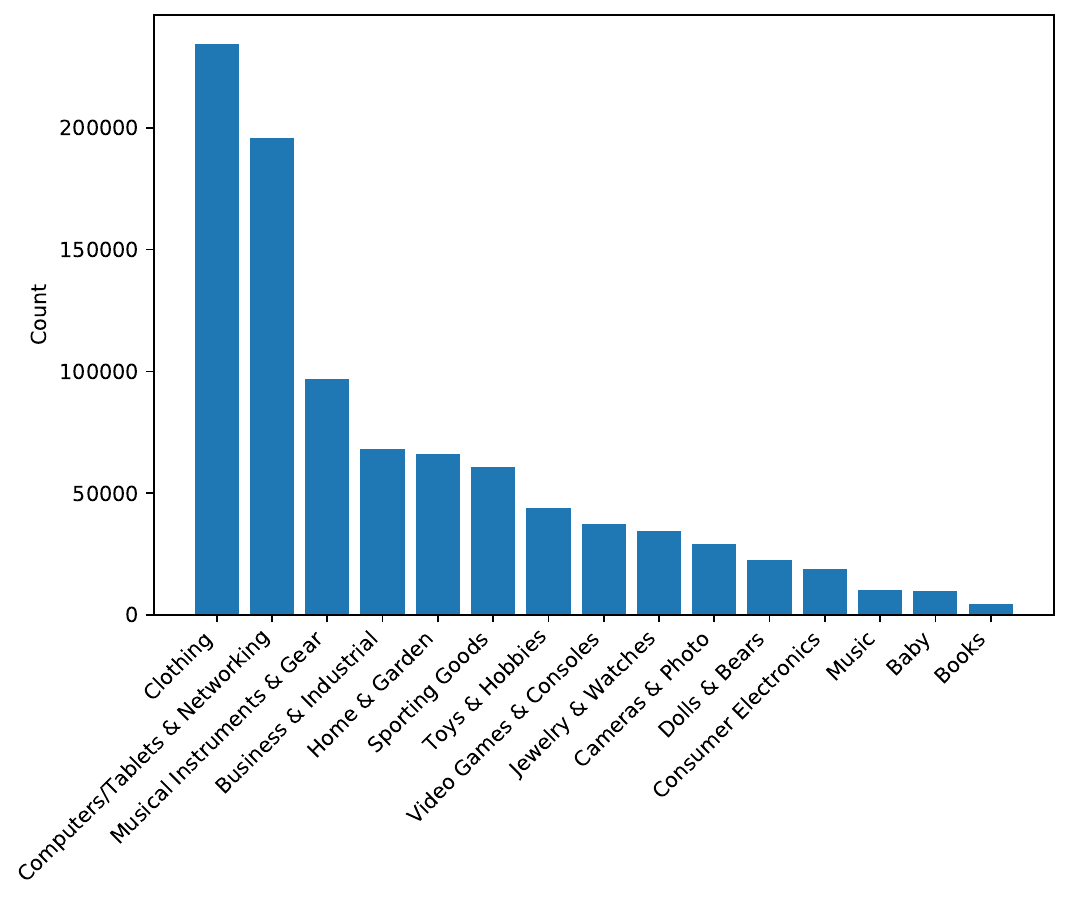}
  \caption{Index set distribution over meta category.}
  \label{fig:meta_dist}
\end{figure}

In this paper, we collect a large-scale image-text online product dataset, ITOP, from an online shopping website \footnote{Unfortunately, we don't plan to release this dataset for unrestricted download.}. ITOP is a large-scale, diverse, well-distributed and cleanly annotated dataset for training, given a large number of likely fine-grained categories, many with subtle differences not easily distinguishable. ITOP has two sets: query set and index set which is the search space for queries. Some examples from the index set are shown in Figure \ref{fig:data_examples}. ITOP uses the hierarchical taxonomy structure which is a sub-tree from a real and large eCommerce inventory. It includes 15 meta categories and 1275 leaf categories. 5,000 query products alongside 1.1 million index products are used for a search benchmark. Its statics are shown in Table \ref{tab:vtop statics}. The sampled 15 meta-category distribution of the index set is shown in Figure \ref{fig:meta_dist}, a long-tailed distribution.

\textbf{Construction of Query Set}
Diversity was prioritized when collecting the queries. Firstly, we collect a 25 million dataset which covers 27 meta and 3,500+ leaf categories. We then manually select over 200 leaf categories from the 25 million dataset to construct our 5,000 queries. The query set is further randomly split by a ratio of 2/3 into a development/test set. 

\textbf{Construction of Index Set} The index set is the search space in our search task. It implicitly consists of two subsets, ground-truth matches and distractors. A ground-truth match is an exact product match to any of the queries, while a distractor is not a match to any queries. A clean and large-scale index set makes a multi-modal search dataset valuable in terms of accurate evaluations of a pre-trained cross-modal model. The index set contains 1.1 million products that cover 1,275 leaf categories. More details on the construction of the index set are shown in Appendix \ref{sec:index set}.

\begin{figure}[t]
  \centering
  \includegraphics[width=0.48\textwidth]{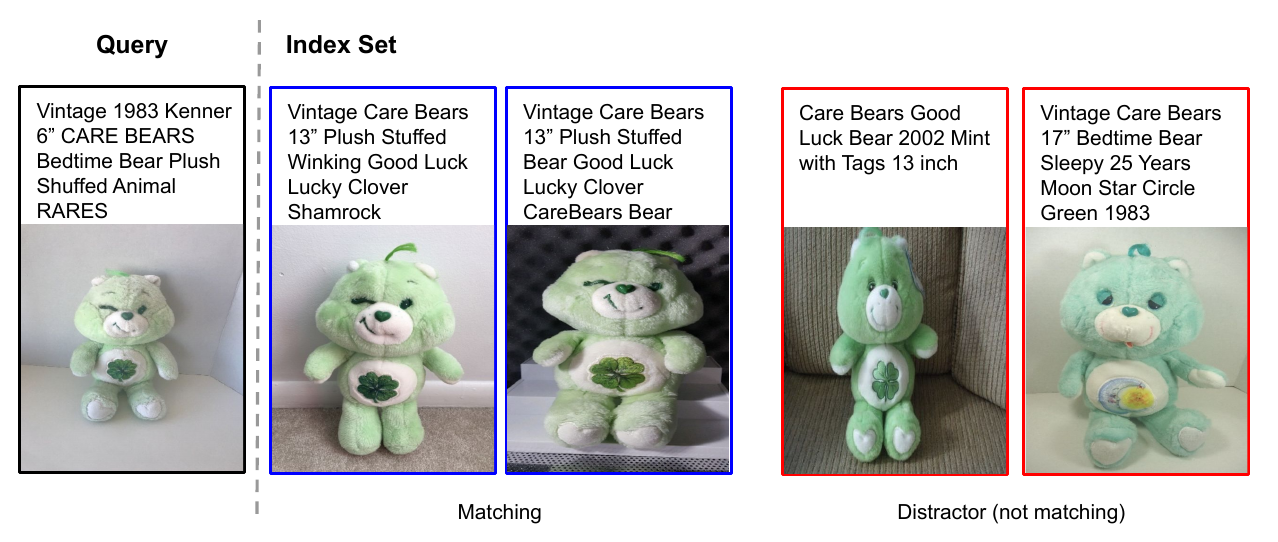}
  \caption{A search example of ITOP: we show a query on the left with its two true matches and two distractors on the right. Distractors are ``hard'' examples because they all come from the same leaf category as the query, i.e. "Care Bears", yet only the true matches share the same product model.}
  \label{fig:match_distractor}
\end{figure}

With all of these procedures, we finally built a large dataset for super fine-grained recognition. A search example is shown in Figure \ref{fig:match_distractor}. 

\begin{table}
  \begin{center}
    \begin{tabular}{c|c|c|c|c} 
      \hline
      Subset & $\#$ of Images & Titles & Meta & Leaf \\
      \hline
      \hline
      Index & 1,101,396 & Yes & 15 & 1,275 \\
      Query & 5,000 & Yes & - & - \\
      \hline
    \end{tabular}
    \caption{ITOP statics. There are not meta or leaf categories for products in the query set.}
    \label{tab:vtop statics}
  \end{center}
\end{table}

\begin{figure*}
  \centering
  \includegraphics[width=0.8\textwidth]{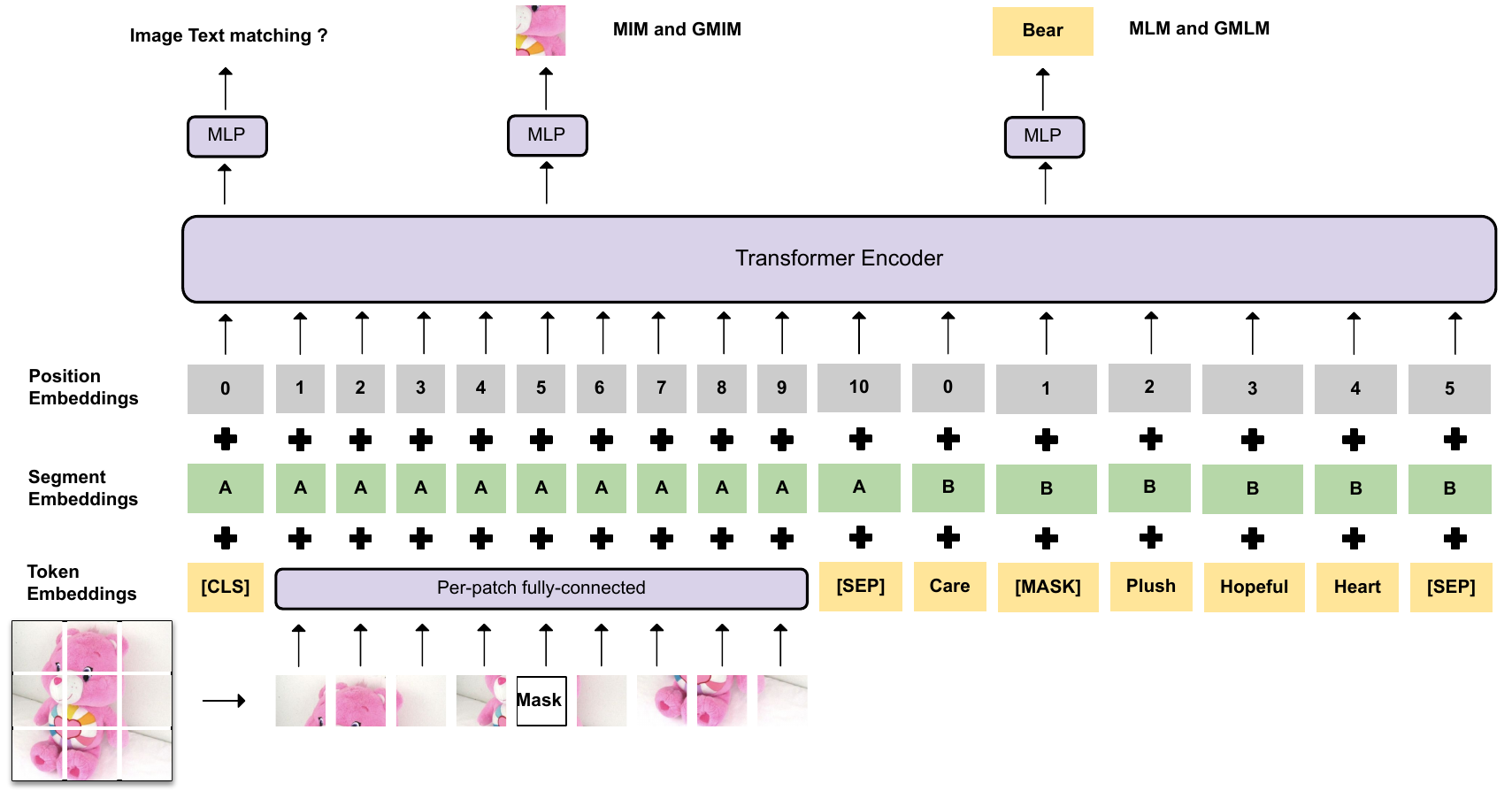}
  \caption{ITEm Architecture. ITEm is pre-trained with five objectives: image-text matching (ITM), masked image modeling (MIM), masked image modeling based on global information (GMIM), masked language modeling (MLM) and masked language modeling based on global information (GMLM). Image patches or tokens are randomly sampled to be masked without knowing regions of interest.}
  \label{fig:arch}
\end{figure*}

\section{Methodology}
Figure \ref{fig:arch} illustrates the ITEm architecture. Similar to BERT \cite{devlin2018bert}, we use a transformer encoder as our basic layer. The image and text are encoded into different embeddings with different embedding layers. Then these embeddings are fed into a bidirectional self-attention transformer encoder to model the cross-modal relationship between image and text. 

ITEm is pre-trained with five unsupervised tasks, trying to learn the relationship between image and text and making the model attend to two modalities as equally as possible.

\textbf{Image Text Matching (ITM)} Some important cross-modal downstream tasks, e.g. category prediction and visual question answering, require the understanding of the relationship between image and text. Similar to the next sentence prediction task of BERT, we pre-train ITEm for a binary matching prediction task. During our pre-training, the product image and title are not always matched. 50$\%$ of the time, the product title is the actual title of the image. 50$\%$ of the time, the product title is a random title from an arbitrary product. The global representation, i.e. the output representation of [CLS] token, is fed into a fully connected layer to obtain the vision-text similarity score. A binary classification loss is used for optimization.

\textbf{Masked Language Modeling (MLM)} MLM is the task of predicting missing tokens in a sequence from their placeholders. Same as BERT's implementation, in 15\% of the title tokens, 80\% are replaced by [MASK], 10\% are replaced with a random token and 10\% are kept unchanged. We build a fully connected layer and a token embedding layer on top of the output representations of these tokens to predict them. A cross-entropy loss is used for optimization.

\textbf{Masked Image Modeling (MIM)} Inspired by \citet{chen2019uniter}, MIM task is to construct the masked patches of an image from their placeholders. In 15\% of the image patches, 80\% are masked and 20\% are kept unchanged. A fully connected layer is built on top of these masked output representations to project them back to the original patch dimension. An L2 loss is applied to regress the original image patches.

\textbf{Masked Language Modeling Based on Global Information (GMLM)} Compared to the next sentence prediction in BERT's implementation, ITM is an easier task. Empirically, we can obtain over 98\% accuracy for ITM within a few iterations. This easy task makes it difficult to summarize the cross-modal information in a global output representation, i.e. the output representation of [CLS] token. 

To summarize both image and title information in the global output representation, we design a new objective, predicting the missing tokens with the global output representation and their corresponding position embeddings rather than their individual output representations. Supposed $t_{n}$ is the masked token, we want to predict the original token with its own position embedding $\bm{p}_n$ and the global output representation $\bm{h}([\rm{CLS}])$. 
\begin{align}
\bm{y}_n = f(\bm{h}(\mathrm{[CLS]}), \bm{p}_n)
\label{equa:representation function}
\end{align}

Inspired by the span boundary objective in \citet{joshi2020spanbert}, we implement the representation function $f(\cdot)$ as a 2-layer feed-forward network with GeLU activation \cite{hendrycks2016gaussian} and layer normalization \cite{ba2016layer}. We use $\bm{y}_n$ to predict the masked token $t_n$. In Figure \ref{fig:arch}, it is $\mathcal{L}_{\mathrm{GMLM}}(\mathrm{Bear})=-\mathrm{log}(\mathrm{Bear}| \bm{h}(\mathrm{[CLS]}), \bm{p}_1)$. With this objective, we force the model to summarize the information of the masked tokens in the global output representation.

\textbf{Masked Image Modeling Based on Global Information (GMIM)} Similar to GMLM on the text side, we design GMIM on the image side to construct the masked patches with the global output representation and their individual position embeddings, forcing the model to summarize the information of the masked patches in the global output representation. Supposed $i_m$ is the masked patch and $\bm{p}_m$ is its position embedding:

\begin{align}
\bm{z}_m = g(\bm{h}(\mathrm{[CLS]}), \bm{p}_m)
\label{equa:representation function}
\end{align}
$g(\cdot)$ shares the same architecture as $f(\cdot)$. We use $\bm{z}_m$ to construct the masked patch.

Combining these five objectives, the final loss for optimization is:
\begin{align}
\mathcal{L}_{\mathrm{total}} =& \lambda_{\mathrm{ITM}} \mathcal{L}_{\mathrm{ITM}} + \lambda_{\mathrm{MLM}} \mathcal{L}_{\mathrm{MLM}} \nonumber\\
    & + \lambda_{\mathrm{GMLM}} \mathcal{L}_{\mathrm{GMLM}} + \lambda_{\mathrm{MIM}} \mathcal{L}_{\mathrm{MIM}} \nonumber \\
    & + \lambda_{\mathrm{GMIM}} \mathcal{L}_{\mathrm{GMIM}}
\label{equa:total loss}
\end{align}
Empirically, we can obtain good performance without tuning the interpolation weight $\lambda$. We only need to set $\lambda_{\rm{ITM}} = 1$, $\lambda_{\rm{MLM}} = \lambda_{\rm{GMLM}} = 0.1$ and $\lambda_{\rm{MIM}} = \lambda_{\rm{GMIM}} = 0.01$, making sure all of these losses are at the same scale.

\section{Experimental Setup}

\subsection{Tasks}
There are two downstream tasks to evaluate our pre-trained ITEm: same product recommendation and leaf category prediction. 

\textbf{Same Product Recommendation} 
As shown in Figure \ref{fig:match_distractor}, this task requires fine-grained embedding to retrieve the same product according to the query. Compared to a similar product recommendation, the same product recommendation has a higher requirement for cross-modal embedding. It requires the embedding to be good enough for subtle differences among products. Since the image is the primary source annotation (see Appendix \ref{sec:index set}), this task also requires the embedding to be good at extracting features from the image, not only from the informatively dominant modality, title. 

For each query, the search space is the whole index set. Even though restricting the search space to the same category as the query is common in eCommerce, this setting makes the task more difficult. Our major criteria for this task are Macro-Average Recall@k (MAR@k) \cite{yang1999evaluation} and Mean Average Precision@k (MAP@k) \cite{manning_raghavan_schuetze_2008} with $k=10$. The higher the better.

\textbf{Leaf Category Prediction} 
Leaf category prediction is another important task in eCommerce. The same product recommendation evaluates the pre-trained model directly, while leaf category prediction requires the pre-trained model to be fine-tuned for this task. Since the query set doesn't contain a leaf category, we don't use it for this task. We split the index set to get the test set (check Section \ref{subsection: implementation} for more details).

\subsection{Baselines}
\textbf{ResNet50} We train ResNet50 with images \cite{DBLP:conf/cvpr/HeZRS16} in a supervised way on the leaf category prediction. The image embedding is obtained before the classifier layer.  

\textbf{RoBERTa} We pre-train RoBERTa$_{\rm{base}}$ \cite{DBLP:journals/corr/abs-1907-11692} with titles. The average embedding of the whole title sequence from the last layer is the final embedding. It performs better than the embedding of [CLS] token.

\textbf{CLIP} We pre-train CLIP \cite{radford2021learning} with both images and titles. Both text encoder and visual encoder share the same setting as BERT$_{\rm{base}}$ encoder. We concatenate the embeddings from the text encoder and visual encoder as the final embedding.

\textbf{Supervised ITEm} We train ITEm with both images and titles in a supervised way on the leaf category prediction. The [CLS] embedding from
the last layer is the final embedding.

No public pre-trained models are applied for fair comparison. All models are trained on ITOP. More implementation details for baseline models are in Appendix \ref{sec:training setting}.

\section{Results and Analysis}
\begin{table*}[ht!]
\centering
\begin{tabular}{l|c|c|c|c}
\hline
 & \multicolumn{2}{c|}{\textbf{Dev. Query Set (\%)}} & \multicolumn{2}{c}{\textbf{Test Query Set (\%)}}\\

\multicolumn{1}{c|}{\textbf{Model}} & \textbf{MAR@10} & \textbf{MAP@10} & \textbf{MAR@10} & \textbf{MAP@10} \\
\hline
\hline
\emph{ResNet50} & 38.51 & 30.67 & 38.43 & 30.24 \\
\emph{RoBERTa} & 42.61 & 37.25 & 46.27 & 41.13 \\
\emph{CLIP} & 59.05 & 52.49 & 57.98 & 51.50 \\
\emph{supervised ITEm} & 54.79 & 49.46 & 54.14 & 48.73 \\
\emph{ITEm} & \textbf{67.53} & \textbf{60.56} & \textbf{66.96} & \textbf{60.05} \\
\hline
\end{tabular}
\caption{\label{tab:search results} 
The Performance of different models on the same product recommendation task. The higher the better.}
\end{table*}

\begin{table*}
\centering
\begin{tabular}{l|l|l|l|l}
\hline
 & \multicolumn{2}{c|}{\textbf{Valid Set (\%)}} & \multicolumn{2}{c}{\textbf{Test Set (\%)}}\\
\multicolumn{1}{c|}{\textbf{Model}} & \textbf{Acc@1} & \textbf{Acc@5} & \textbf{Acc@1} & \textbf{Acc@5} \\
\hline
\hline
\emph{ResNet50} & 74.72 & 90.68 & 74.92 & 90.84 \\
\emph{RoBERTa} & 88.79 & 98.07 & 88.92 & 98.05 \\
\emph{CLIP} & 90.11 & 98.16 & 90.42 & 98.21 \\
\emph{supervised ITEm} & 89.49 & 98.12 & 89.82 & 98.11 \\
\emph{ITEm} & \textbf{90.79} & \textbf{98.51} & \textbf{91.09} & \textbf{98.52} \\
\hline
\end{tabular}
\caption{\label{tab: leaf category} 
The Accuracy of different models on the leaf category prediction task.}
\end{table*}
\subsection{Implementation}
\label{subsection: implementation}
The index set is randomly split by a ratio of 9/1 into train/non-train sets. All pre-training and fine-tuning are conducted on the train set. The non-train set is further randomly split by a ratio of 1/1 into a valid/test set for leaf category prediction.

We implement ITEm in fairseq \cite{ott2019fairseq}. For the pre-training, the learning rate is warmed up over the first 4,000 steps to a peak value of 1e-4, and then linearly decayed. We set $\beta$ hyperparameters ($\beta_1=0.9$, $\beta_2=0.98$) and a decoupled weight decay \cite{loshchilov2017decoupled} of 1e-4. We deviate from the optimization by running for 100k steps and using an $\epsilon$ of 1e-8 for AdamW \cite{kingma2014adam}. 

The ITEm architecture shares the same setting as $\rm{BERT_{base}}$. We use a batch size of 512, a max title length of 36, and an image size of 224 with 16 as the patch size. Random resizing and horizontal flip techniques are used for image augmentation. We build our own BPE \cite{DBLP:conf/acl/SennrichHB16a} vocabulary with a size of 50K.

We employ cosine distance to measure the similarity between a query and an index product:
\begin{align}
s_g(i, j) = (\bm{e}_g^q)^T \bm{e}_g^i
\label{equa:cosine similarity}
\end{align}
where $\bm{e}_g^q$ is the [CLS] (global) embedding for the q-th query and $\bm{e}_g^i$ is the global embedding for the i-th index product. The similarity score for the text embedding $s_t(i, j)$ (average embedding over the whole title sequence) or the vision embedding $s_v(i, j)$ (average embedding over the whole image sequence) have a similar definition. The final similarity score for the q-th query and the i-th index product is defined as:
\begin{align}
s(i, j) = \mathrm{max}(s_g(i, j), s_t(i, j), s_v(i, j))
\label{equa:cosine similarity}
\end{align}

The pre-training was done on 8 Volta V100 GPUs and took 2 days to complete. The best checkpoint is determined by the minimal $\mathcal{L}_{\rm{total}}$ on the valid set. All results are reported by the mean of at least three trials.

\subsection{Same Product Recommendation}
As shown in Table \ref{tab:search results}, the cross-modal models (ITEm, supervised ITEm and CLIP) outperform other models trained on a single modality (RoBERTa and ResNet50), because complementary information is offered from different modalities. Another observation is that the models trained with unsupervised learning achieve better results than the ones with supervised learning. When training to predict the leaf category, in a supervised way, the model tends to learn the common pattern within each category. During the search, the supervised-learning model might consider the products with the same leaf category as similar products, resulting in a worse score for the same product recommendation task. The best result is obtained by ITEm. It is powerful enough to retrieve the products with similar images and similar titles, resulting in the same products. Some retrieval examples are shown in Appendix \ref{sec:query index}.

\subsection{Leaf Category Prediction}
Table \ref{tab: leaf category} shows leaf category prediction results from different methods. ITEm with a pre-training procedure achieves the best result, which shows that our pre-training objectives help gather the information from different modalities. In addition, there is a huge performance gap between ResNet50 and RoBERTa, with only a subtle difference between RoBERTa and other cross-modal models. This is caused by the bias of our ITOP dataset, where the title generally plays the major role and contains more information than the image for most products. In some cases, the leaf category names could even appear in the titles.

\section{Conclusion}
In this paper, we present ITEm, a pre-trained model for joint image-and-text representation. With five pre-training objectives, especially the two newly designed objectives that lead ITEm to extract information from images and text more equally, and induce ITEm to learn fine-grained features for the product. The embedding from a pre-trained ITEm is used for the search for extremely similar products. We also fine-tuned ITEm on the prediction task and obtained very good accuracy. 

In the future, we plan to evaluate our model on public datasets that are rare at this moment. If possible, we want to make parts of our own collected dataset public or provide it for research usage.

\section{Ethical Considerations}
We collect anonymous user data in full compliance with existing legislation (e.g., GDPR). We aim to create a large-scale, diverse, well-distributed, and cleanly annotated dataset for training given a large number of likely fine-grained categories, many with subtle differences not easily distinguishable. Similar to many other fine-grained datasets, a long-tailed distribution is typical in our dataset. However, we have clipped the top of the largest categories to reduce their dominance. Due to the clipping process, the data distribution is re-balanced by restricting the super-large categories and resulting in raising super-tiny ones. The re-balancing is also motivated to build a genuinely equitable marketplace for the world to buy and sell diverse items.

\bibliography{custom}

\begin{thebibliography}{40}
\expandafter\ifx\csname natexlab\endcsname\relax\def\natexlab#1{#1}\fi

\bibitem[{Antol et~al.(2015)Antol, Agrawal, Lu, Mitchell, Batra, Zitnick, and Parikh}]{antol2015vqa}
Stanislaw Antol, Aishwarya Agrawal, Jiasen Lu, Margaret Mitchell, Dhruv Batra, C.~Lawrence Zitnick, and Devi Parikh. 2015.
\newblock \href {https://doi.org/10.1109/ICCV.2015.279} {Vqa: Visual question answering}.
\newblock In \emph{2015 IEEE International Conference on Computer Vision (ICCV)}, pages 2425--2433.

\bibitem[{Ba et~al.(2016)Ba, Kiros, and Hinton}]{ba2016layer}
Lei~Jimmy Ba, Jamie~Ryan Kiros, and Geoffrey~E. Hinton. 2016.
\newblock \href {http://arxiv.org/abs/1607.06450} {Layer normalization}.
\newblock \emph{CoRR}, abs/1607.06450.

\bibitem[{Bianchi et~al.(2021)Bianchi, Tagliabue, and Yu}]{bianchi2021query2prod2vec}
Federico Bianchi, Jacopo Tagliabue, and Bingqing Yu. 2021.
\newblock \href {https://doi.org/10.18653/v1/2021.naacl-industry.20} {Query2prod2vec: Grounded word embeddings for ecommerce}.
\newblock In \emph{Proceedings of the 2021 Conference of the North American Chapter of the Association for Computational Linguistics: Human Language Technologies: Industry Papers, {NAACL-HLT} 2021, Online, June 6-11, 2021}, pages 154--162. Association for Computational Linguistics.

\bibitem[{Chen et~al.(2015)Chen, Fang, Lin, Vedantam, Gupta, Doll{\'{a}}r, and Zitnick}]{chen2015microsoft}
Xinlei Chen, Hao Fang, Tsung{-}Yi Lin, Ramakrishna Vedantam, Saurabh Gupta, Piotr Doll{\'{a}}r, and C.~Lawrence Zitnick. 2015.
\newblock \href {http://arxiv.org/abs/1504.00325} {Microsoft {COCO} captions: Data collection and evaluation server}.
\newblock \emph{CoRR}, abs/1504.00325.

\bibitem[{Chen et~al.(2019)Chen, Li, Yu, Kholy, Ahmed, Gan, Cheng, and Liu}]{chen2019uniter}
Yen{-}Chun Chen, Linjie Li, Licheng Yu, Ahmed~El Kholy, Faisal Ahmed, Zhe Gan, Yu~Cheng, and Jingjing Liu. 2019.
\newblock \href {http://arxiv.org/abs/1909.11740} {{UNITER:} learning universal image-text representations}.
\newblock \emph{CoRR}, abs/1909.11740.

\bibitem[{Desai and Johnson(2021)}]{desai2020virtex}
Karan Desai and Justin Johnson. 2021.
\newblock \href {https://openaccess.thecvf.com/content/CVPR2021/html/Desai\_VirTex\_Learning\_Visual\_Representations\_From\_Textual\_Annotations\_CVPR\_2021\_paper.html} {Virtex: Learning visual representations from textual annotations}.
\newblock In \emph{{IEEE} Conference on Computer Vision and Pattern Recognition, {CVPR} 2021, virtual, June 19-25, 2021}, pages 11162--11173. Computer Vision Foundation / {IEEE}.

\bibitem[{Devlin et~al.(2019)Devlin, Chang, Lee, and Toutanova}]{devlin2018bert}
Jacob Devlin, Ming{-}Wei Chang, Kenton Lee, and Kristina Toutanova. 2019.
\newblock \href {https://doi.org/10.18653/v1/n19-1423} {{BERT:} pre-training of deep bidirectional transformers for language understanding}.
\newblock In \emph{Proceedings of the 2019 Conference of the North American Chapter of the Association for Computational Linguistics: Human Language Technologies, {NAACL-HLT} 2019, Minneapolis, MN, USA, June 2-7, 2019, Volume 1 (Long and Short Papers)}, pages 4171--4186. Association for Computational Linguistics.

\bibitem[{Dosovitskiy et~al.(2021)Dosovitskiy, Beyer, Kolesnikov, Weissenborn, Zhai, Unterthiner, Dehghani, Minderer, Heigold, Gelly, Uszkoreit, and Houlsby}]{dosovitskiy2020image}
Alexey Dosovitskiy, Lucas Beyer, Alexander Kolesnikov, Dirk Weissenborn, Xiaohua Zhai, Thomas Unterthiner, Mostafa Dehghani, Matthias Minderer, Georg Heigold, Sylvain Gelly, Jakob Uszkoreit, and Neil Houlsby. 2021.
\newblock \href {https://openreview.net/forum?id=YicbFdNTTy} {An image is worth 16x16 words: Transformers for image recognition at scale}.
\newblock In \emph{9th International Conference on Learning Representations, {ICLR} 2021, Virtual Event, Austria, May 3-7, 2021}. OpenReview.net.

\bibitem[{Goyal et~al.(2019)Goyal, Khot, Agrawal, Summers{-}Stay, Batra, and Parikh}]{goyal2017making}
Yash Goyal, Tejas Khot, Aishwarya Agrawal, Douglas Summers{-}Stay, Dhruv Batra, and Devi Parikh. 2019.
\newblock \href {https://doi.org/10.1007/s11263-018-1116-0} {Making the {V} in {VQA} matter: Elevating the role of image understanding in visual question answering}.
\newblock \emph{Int. J. Comput. Vis.}, 127(4):398--414.

\bibitem[{Grbovic et~al.(2015)Grbovic, Radosavljevic, Djuric, Bhamidipati, Savla, Bhagwan, and Sharp}]{10.1145/2783258.2788627}
Mihajlo Grbovic, Vladan Radosavljevic, Nemanja Djuric, Narayan Bhamidipati, Jaikit Savla, Varun Bhagwan, and Doug Sharp. 2015.
\newblock \href {https://doi.org/10.1145/2783258.2788627} {E-commerce in your inbox: Product recommendations at scale}.
\newblock In \emph{Proceedings of the 21th ACM SIGKDD International Conference on Knowledge Discovery and Data Mining}, KDD '15, page 1809–1818, New York, NY, USA. Association for Computing Machinery.

\bibitem[{He et~al.(2016)He, Zhang, Ren, and Sun}]{DBLP:conf/cvpr/HeZRS16}
Kaiming He, Xiangyu Zhang, Shaoqing Ren, and Jian Sun. 2016.
\newblock \href {https://doi.org/10.1109/CVPR.2016.90} {Deep residual learning for image recognition}.
\newblock In \emph{2016 {IEEE} Conference on Computer Vision and Pattern Recognition, {CVPR} 2016, Las Vegas, NV, USA, June 27-30, 2016}, pages 770--778. {IEEE} Computer Society.

\bibitem[{Hendrycks and Gimpel(2016)}]{hendrycks2016gaussian}
Dan Hendrycks and Kevin Gimpel. 2016.
\newblock \href {http://arxiv.org/abs/1606.08415} {Bridging nonlinearities and stochastic regularizers with gaussian error linear units}.
\newblock \emph{CoRR}, abs/1606.08415.

\bibitem[{Jing et~al.(2015)Jing, Liu, Kislyuk, Zhai, Xu, Donahue, and Tavel}]{jing2015visual}
Yushi Jing, David~C. Liu, Dmitry Kislyuk, Andrew Zhai, Jiajing Xu, Jeff Donahue, and Sarah Tavel. 2015.
\newblock \href {https://doi.org/10.1145/2783258.2788621} {Visual search at pinterest}.
\newblock In \emph{Proceedings of the 21th {ACM} {SIGKDD} International Conference on Knowledge Discovery and Data Mining, Sydney, NSW, Australia, August 10-13, 2015}, pages 1889--1898. {ACM}.

\bibitem[{Joshi et~al.(2020)Joshi, Chen, Liu, Weld, Zettlemoyer, and Levy}]{joshi2020spanbert}
Mandar Joshi, Danqi Chen, Yinhan Liu, Daniel~S. Weld, Luke Zettlemoyer, and Omer Levy. 2020.
\newblock \href {https://transacl.org/ojs/index.php/tacl/article/view/1853} {Spanbert: Improving pre-training by representing and predicting spans}.
\newblock \emph{Trans. Assoc. Comput. Linguistics}, 8:64--77.

\bibitem[{Kazemzadeh et~al.(2014)Kazemzadeh, Ordonez, Matten, and Berg}]{kazemzadeh2014referitgame}
Sahar Kazemzadeh, Vicente Ordonez, Mark Matten, and Tamara Berg. 2014.
\newblock \href {https://doi.org/10.3115/v1/D14-1086} {{R}efer{I}t{G}ame: Referring to objects in photographs of natural scenes}.
\newblock In \emph{Proceedings of the 2014 Conference on Empirical Methods in Natural Language Processing ({EMNLP})}, pages 787--798, Doha, Qatar. Association for Computational Linguistics.

\bibitem[{Keenan(2021)}]{michkee}
Michael Keenan. 2021.
\newblock \href {https://www.shopify.com/enterprise/global-ecommerce-statistics} {Global ecommerce explained: Stats and trends to watch in 2021}.

\bibitem[{Kiefer and Wolfowitz(1952)}]{kiefer1952stochastic}
J.~Kiefer and J.~Wolfowitz. 1952.
\newblock \href {https://doi.org/10.1214/aoms/1177729392} {Stochastic estimation of the maximum of a regression function}.
\newblock \emph{The Annals of Mathematical Statistics}, 22(3):462--466.

\bibitem[{Kingma and Ba(2015)}]{kingma2014adam}
Diederik~P. Kingma and Jimmy Ba. 2015.
\newblock \href {http://arxiv.org/abs/1412.6980} {Adam: {A} method for stochastic optimization}.
\newblock In \emph{3rd International Conference on Learning Representations, {ICLR} 2015, San Diego, CA, USA, May 7-9, 2015, Conference Track Proceedings}.

\bibitem[{Li et~al.(2019)Li, Yatskar, Yin, Hsieh, and Chang}]{li2019visualbert}
Liunian~Harold Li, Mark Yatskar, Da~Yin, Cho{-}Jui Hsieh, and Kai{-}Wei Chang. 2019.
\newblock \href {http://arxiv.org/abs/1908.03557} {Visualbert: {A} simple and performant baseline for vision and language}.
\newblock \emph{CoRR}, abs/1908.03557.

\bibitem[{Liu et~al.(2019)Liu, Ott, Goyal, Du, Joshi, Chen, Levy, Lewis, Zettlemoyer, and Stoyanov}]{DBLP:journals/corr/abs-1907-11692}
Yinhan Liu, Myle Ott, Naman Goyal, Jingfei Du, Mandar Joshi, Danqi Chen, Omer Levy, Mike Lewis, Luke Zettlemoyer, and Veselin Stoyanov. 2019.
\newblock \href {http://arxiv.org/abs/1907.11692} {Roberta: {A} robustly optimized {BERT} pretraining approach}.
\newblock \emph{CoRR}, abs/1907.11692.

\bibitem[{Loshchilov and Hutter(2019)}]{loshchilov2017decoupled}
Ilya Loshchilov and Frank Hutter. 2019.
\newblock \href {https://openreview.net/forum?id=Bkg6RiCqY7} {Decoupled weight decay regularization}.
\newblock In \emph{7th International Conference on Learning Representations, {ICLR} 2019, New Orleans, LA, USA, May 6-9, 2019}. OpenReview.net.

\bibitem[{Lu et~al.(2019)Lu, Batra, Parikh, and Lee}]{lu2019vilbert}
Jiasen Lu, Dhruv Batra, Devi Parikh, and Stefan Lee. 2019.
\newblock \href {https://proceedings.neurips.cc/paper/2019/hash/c74d97b01eae257e44aa9d5bade97baf-Abstract.html} {Vilbert: Pretraining task-agnostic visiolinguistic representations for vision-and-language tasks}.
\newblock In \emph{Advances in Neural Information Processing Systems 32: Annual Conference on Neural Information Processing Systems 2019, NeurIPS 2019, December 8-14, 2019, Vancouver, BC, Canada}, pages 13--23.

\bibitem[{Manning et~al.(2008)Manning, Raghavan, and Schütze}]{manning_raghavan_schuetze_2008}
Christopher~D. Manning, Prabhakar Raghavan, and Hinrich Schütze. 2008.
\newblock \href {https://doi.org/10.1017/CBO9780511809071.009} {\emph{Evaluation in information retrieval}}, page 139–161. Cambridge University Press.

\bibitem[{Ott et~al.(2019)Ott, Edunov, Baevski, Fan, Gross, Ng, Grangier, and Auli}]{ott2019fairseq}
Myle Ott, Sergey Edunov, Alexei Baevski, Angela Fan, Sam Gross, Nathan Ng, David Grangier, and Michael Auli. 2019.
\newblock \href {https://doi.org/10.18653/v1/n19-4009} {fairseq: {A} fast, extensible toolkit for sequence modeling}.
\newblock In \emph{Proceedings of the 2019 Conference of the North American Chapter of the Association for Computational Linguistics: Human Language Technologies, {NAACL-HLT} 2019, Minneapolis, MN, USA, June 2-7, 2019, Demonstrations}, pages 48--53. Association for Computational Linguistics.

\bibitem[{Periyathambi et~al.(2023)Periyathambi, Lancewicki, Hewavitharana, Reeves, Padmanabhan, Stein, Boorn, Liao, Becker, Elkis et~al.}]{periyathambi2023systems}
Ramesh Periyathambi, Tomer Lancewicki, Sanjika Hewavitharana, Ryan Reeves, Senthil~Kumar Padmanabhan, Daniel Stein, Luther~S Boorn, Baohao Liao, Simon~Alexander Becker, Sivan Elkis, et~al. 2023.
\newblock Systems and methods for creating listing for items for sale in an electronic marketplace.
\newblock US Patent App. 17/562,423.

\bibitem[{Plummer et~al.(2015)Plummer, Wang, Cervantes, Caicedo, Hockenmaier, and Lazebnik}]{plummer2015flickr30k}
Bryan~A. Plummer, Liwei Wang, Chris~M. Cervantes, Juan~C. Caicedo, Julia Hockenmaier, and Svetlana Lazebnik. 2015.
\newblock \href {https://doi.org/10.1109/ICCV.2015.303} {Flickr30k entities: Collecting region-to-phrase correspondences for richer image-to-sentence models}.
\newblock In \emph{2015 {IEEE} International Conference on Computer Vision, {ICCV} 2015, Santiago, Chile, December 7-13, 2015}, pages 2641--2649. {IEEE} Computer Society.

\bibitem[{Qi et~al.(2020)Qi, Su, Song, Cui, Bharti, and Sacheti}]{qi2020imagebert}
Di~Qi, Lin Su, Jia Song, Edward Cui, Taroon Bharti, and Arun Sacheti. 2020.
\newblock \href {http://arxiv.org/abs/2001.07966} {Imagebert: Cross-modal pre-training with large-scale weak-supervised image-text data}.
\newblock \emph{CoRR}, abs/2001.07966.

\bibitem[{Radford et~al.(2021)Radford, Kim, Hallacy, Ramesh, Goh, Agarwal, Sastry, Askell, Mishkin, Clark, Krueger, and Sutskever}]{radford2021learning}
Alec Radford, Jong~Wook Kim, Chris Hallacy, Aditya Ramesh, Gabriel Goh, Sandhini Agarwal, Girish Sastry, Amanda Askell, Pamela Mishkin, Jack Clark, Gretchen Krueger, and Ilya Sutskever. 2021.
\newblock \href {http://proceedings.mlr.press/v139/radford21a.html} {Learning transferable visual models from natural language supervision}.
\newblock In \emph{Proceedings of the 38th International Conference on Machine Learning, {ICML} 2021, 18-24 July 2021, Virtual Event}, volume 139 of \emph{Proceedings of Machine Learning Research}, pages 8748--8763. {PMLR}.

\bibitem[{Sennrich et~al.(2016)Sennrich, Haddow, and Birch}]{DBLP:conf/acl/SennrichHB16a}
Rico Sennrich, Barry Haddow, and Alexandra Birch. 2016.
\newblock \href {https://doi.org/10.18653/v1/p16-1162} {Neural machine translation of rare words with subword units}.
\newblock In \emph{Proceedings of the 54th Annual Meeting of the Association for Computational Linguistics, {ACL} 2016, August 7-12, 2016, Berlin, Germany, Volume 1: Long Papers}. The Association for Computer Linguistics.

\bibitem[{Shankar et~al.(2017)Shankar, Narumanchi, Ananya, Kompalli, and Chaudhury}]{shankar2017deep}
Devashish Shankar, Sujay Narumanchi, H.~A. Ananya, Pramod Kompalli, and Krishnendu Chaudhury. 2017.
\newblock \href {http://arxiv.org/abs/1703.02344} {Deep learning based large scale visual recommendation and search for e-commerce}.
\newblock \emph{CoRR}, abs/1703.02344.

\bibitem[{Sharma et~al.(2018)Sharma, Ding, Goodman, and Soricut}]{sharma2018conceptual}
Piyush Sharma, Nan Ding, Sebastian Goodman, and Radu Soricut. 2018.
\newblock \href {https://aclanthology.org/P18-1238.pdf} {Conceptual captions: A cleaned, hypernymed, image alt-text dataset for automatic image captioning}.
\newblock In \emph{Proceedings of the 56th Annual Meeting of the Association for Computational Linguistics (Volume 1: Long Papers)}, pages 2556--2565.

\bibitem[{Song et~al.(2016)Song, Xiang, Jegelka, and Savarese}]{oh2016deep}
Hyun~Oh Song, Yu~Xiang, Stefanie Jegelka, and Silvio Savarese. 2016.
\newblock \href {https://doi.org/10.1109/CVPR.2016.434} {Deep metric learning via lifted structured feature embedding}.
\newblock In \emph{2016 {IEEE} Conference on Computer Vision and Pattern Recognition, {CVPR} 2016, Las Vegas, NV, USA, June 27-30, 2016}, pages 4004--4012. {IEEE} Computer Society.

\bibitem[{Su et~al.(2020)Su, Zhu, Cao, Li, Lu, Wei, and Dai}]{su2019vl}
Weijie Su, Xizhou Zhu, Yue Cao, Bin Li, Lewei Lu, Furu Wei, and Jifeng Dai. 2020.
\newblock \href {https://openreview.net/forum?id=SygXPaEYvH} {{VL-BERT:} pre-training of generic visual-linguistic representations}.
\newblock In \emph{8th International Conference on Learning Representations, {ICLR} 2020, Addis Ababa, Ethiopia, April 26-30, 2020}. OpenReview.net.

\bibitem[{Suhr et~al.(2019)Suhr, Zhou, Zhang, Zhang, Bai, and Artzi}]{suhr2018corpus}
Alane Suhr, Stephanie Zhou, Ally Zhang, Iris Zhang, Huajun Bai, and Yoav Artzi. 2019.
\newblock \href {https://doi.org/10.18653/v1/p19-1644} {A corpus for reasoning about natural language grounded in photographs}.
\newblock In \emph{Proceedings of the 57th Conference of the Association for Computational Linguistics, {ACL} 2019, Florence, Italy, July 28- August 2, 2019, Volume 1: Long Papers}, pages 6418--6428. Association for Computational Linguistics.

\bibitem[{Tan and Bansal(2019)}]{tan2019lxmert}
Hao Tan and Mohit Bansal. 2019.
\newblock \href {https://doi.org/10.18653/v1/D19-1514} {{LXMERT:} learning cross-modality encoder representations from transformers}.
\newblock In \emph{Proceedings of the 2019 Conference on Empirical Methods in Natural Language Processing and the 9th International Joint Conference on Natural Language Processing, {EMNLP-IJCNLP} 2019, Hong Kong, China, November 3-7, 2019}, pages 5099--5110. Association for Computational Linguistics.

\bibitem[{Vaswani et~al.(2017)Vaswani, Shazeer, Parmar, Uszkoreit, Jones, Gomez, Kaiser, and Polosukhin}]{vaswani2017attention}
Ashish Vaswani, Noam Shazeer, Niki Parmar, Jakob Uszkoreit, Llion Jones, Aidan~N. Gomez, Lukasz Kaiser, and Illia Polosukhin. 2017.
\newblock \href {https://proceedings.neurips.cc/paper/2017/hash/3f5ee243547dee91fbd053c1c4a845aa-Abstract.html} {Attention is all you need}.
\newblock In \emph{Advances in Neural Information Processing Systems 30: Annual Conference on Neural Information Processing Systems 2017, December 4-9, 2017, Long Beach, CA, {USA}}, pages 5998--6008.

\bibitem[{Yang et~al.(2017)Yang, Kale, Bubnov, Stein, Wang, Kiapour, and Piramuthu}]{yang2017visual}
Fan Yang, Ajinkya Kale, Yury Bubnov, Leon Stein, Qiaosong Wang, M.~Hadi Kiapour, and Robinson Piramuthu. 2017.
\newblock \href {https://doi.org/10.1145/3097983.3098162} {Visual search at ebay}.
\newblock In \emph{Proceedings of the 23rd {ACM} {SIGKDD} International Conference on Knowledge Discovery and Data Mining, Halifax, NS, Canada, August 13 - 17, 2017}, pages 2101--2110. {ACM}.

\bibitem[{Yang(1999)}]{yang1999evaluation}
Yiming Yang. 1999.
\newblock \href {https://doi.org/https://doi.org/10.1023/A:1009982220290} {An evaluation of statistical approaches to text categorization}.
\newblock \emph{Information retrieval}, 1(1):69--90.

\bibitem[{You et~al.(2016)You, Jin, Wang, Fang, and Luo}]{you2016image}
Quanzeng You, Hailin Jin, Zhaowen Wang, Chen Fang, and Jiebo Luo. 2016.
\newblock \href {https://doi.org/10.1109/CVPR.2016.503} {Image captioning with semantic attention}.
\newblock In \emph{2016 {IEEE} Conference on Computer Vision and Pattern Recognition, {CVPR} 2016, Las Vegas, NV, USA, June 27-30, 2016}, pages 4651--4659. {IEEE} Computer Society.

\bibitem[{Zellers et~al.(2019)Zellers, Bisk, Farhadi, and Choi}]{zellers2019recognition}
Rowan Zellers, Yonatan Bisk, Ali Farhadi, and Yejin Choi. 2019.
\newblock \href {https://doi.org/10.1109/CVPR.2019.00688} {From recognition to cognition: Visual commonsense reasoning}.
\newblock In \emph{{IEEE} Conference on Computer Vision and Pattern Recognition, {CVPR} 2019, Long Beach, CA, USA, June 16-20, 2019}, pages 6720--6731. Computer Vision Foundation / {IEEE}.

\end{thebibliography}

\clearpage
\appendix

\section{Construction Details of Index Set}
\label{sec:index set}

To collect both matches and distractors, we first create a candidate pool by randomly sampling from the dataset with 25 million products, resulting in about 1.1 million products that cover 1,275 leaf categories. Next, we need to annotate all candidates to be either an ``exact match'' or ``not-a-match''. To reduce the cost caused by complete human labeling, we used a pre-ranking methodology to reduce the size of shortlists for human review. We also employ categorical filtering, similarity-based thresholding, and other methods to further improve the recalls. During the crow-sourced annotation process, given query-candidate pairs, human annotation review both the image and title and give several levels of confidence for "exact match". Each query-candidate pair has been rated by at most ten annotators. We performed post-processing to determine the final acceptable labels and discarded any labels with low confidence as either "match" or "not match" to minimize the incorrect annotation risk. 

\textbf{Same Product} To define two images as including the ``same product'' is often subjective and challenging especially when critical aspects are missing from their titles and are invisible from images. In our case, we use images as the primary source for annotations. The criteria for ``same product'' in two product images are:
\begin{itemize}
	\item The two products can represent different product conditions. For example, a broken phone and a new phone with the exact same specifications (make, model, color, etc.) are the ``same products''.
	\item They should have the same color/model/style, and other aspects that are visually visible and distinguishable. For example, a golden and a gray phone of otherwise the same make and model are not considered the ``same'' products.
	\item They can vary in other aspects that are not distinguished solely based on image, e.g. shoe size, memory size for hardware, etc.
\end{itemize}

\section{Implementation Details}
\label{sec:training setting}
\textbf{Data Split} The index set is randomly split into train/valid/test set by a ratio of 9/0.5/0.5 (same as Section \ref{subsection: implementation}). We conduct the pre-training of any model on the train set and then fine-tune it on the same train set for the leaf category prediction task. If we don't pre-train the model (like ResNet50 or supervised ITEm), we train it on the train set from scratch. This data split is intentionally designed for a fair comparison between supervised and unsupervised learning methods. 

RoBERTa, CLIP and ITEm are pre-trained first on the train set. We then evaluate their embeddings directly on the same product recommendation. For the leaf category prediction task, the pre-trained model is continuously fine-tuned on the train set. The best checkpoint is determined by the best prediction score on the valid set. ResNet50 and supervised ITEm are only trained once on the train set for the leaf category prediction task.

\textbf{ResNet50} We train ResNet50 in a supervised way on leaf category prediction with an image as input. The initial learning rate is set to 0.1, then decayed by a factor of 0.1 if the performance on the valid set hasn't improved over 5 epochs. We use SGD \cite{kiefer1952stochastic} with a momentum of 0.9. The image size is 224, with random resizing and horizontal flip as the data augmentation methods. The training is stopped when the performance on the valid set hasn't improved over 10 epochs. The batch size is 256.

\textbf{RoBERTa} We pre-train RoBERTa in an unsupervised way with the title as input. The architecture setting stays the same as the original $\rm{BERT_{base}}$ and uses the same vocabulary as ITEm. Same as ITEm, the maximum title length is set to 36. The initial learning rate is warmed up over the first 4000 steps to a peak value of $1e-4$, and then decayed proportionally to the inverse square root of the step number \cite{vaswani2017attention}. We set $\beta$ hyperparameters ($\beta_1= 0.9$, $\beta_2= 0.98$) and decoupled weight decay of 1e-4.  We deviate from the optimization for 100k steps by using an $\epsilon$ of 1e-8 for AdamW. We use a batch size of 512.

For the fine-tuning on the leaf category prediction task, we add an extra classifier on top of the [CLS] embedding from the last layer, setting linear learning rate to 1e-5 with linear rate decay, AdamW with $\beta_1= 0.9$, $\beta_2= 0.98$ and weight decay of 0.1, warmup ratio of 0.06, batch size of 64 and maximum epochs of 10 (same setting as GLUE task in \citet{DBLP:journals/corr/abs-1907-11692}). 

\textbf{CLIP} We pre-train CLIP with both image and title as input. The vocabulary is borrowed from ITEm and the maximum title length is also set to 36. We borrow the training setting from the original paper, \citet{radford2021learning}, except for setting the batch size to 512.

For the fine-tuning of the leaf category prediction task, we concatenate the output [CLS] embeddings from both text and visual encoders and add an extra classifier on top of it. The fine-tuning setting stays the same as RoBERTa.

\textbf{Supervised ITEm} We train ITEm with both image and title as input for leaf category prediction. An extra classifier layer is added on top of the [CLS] embedding from the last layer. The initial learning rate is warmed up over the first 1000 steps to a peak value of $1e-4$, and then decayed proportionally to the inverse square root of the step number \cite{vaswani2017attention}. We set $\beta$ hyperparameters ($\beta_1= 0.9$, $\beta_2= 0.98$) and decoupled weight decay of 1e-4.  We apply AdamW with an $\epsilon$ of 1e-8. The training is stopped when the performance on the valid set hasn't improved over 10 epochs. We use a batch size of 256.

\textbf{Fine-tune ITEm} For the leaf category prediction task, the pre-trained ITEm is continuously fine-tuned on the train set. Same as supervised ITEm, we add an extra classifier layer to the top of [CLS] embedding from the last layer. The training setting stays the same as RoBERTa.

\section{Examples for Same Product Recommendation}
\label{sec:query index}
\begin{figure*}
  \centering
  \includegraphics[width=1\textwidth]{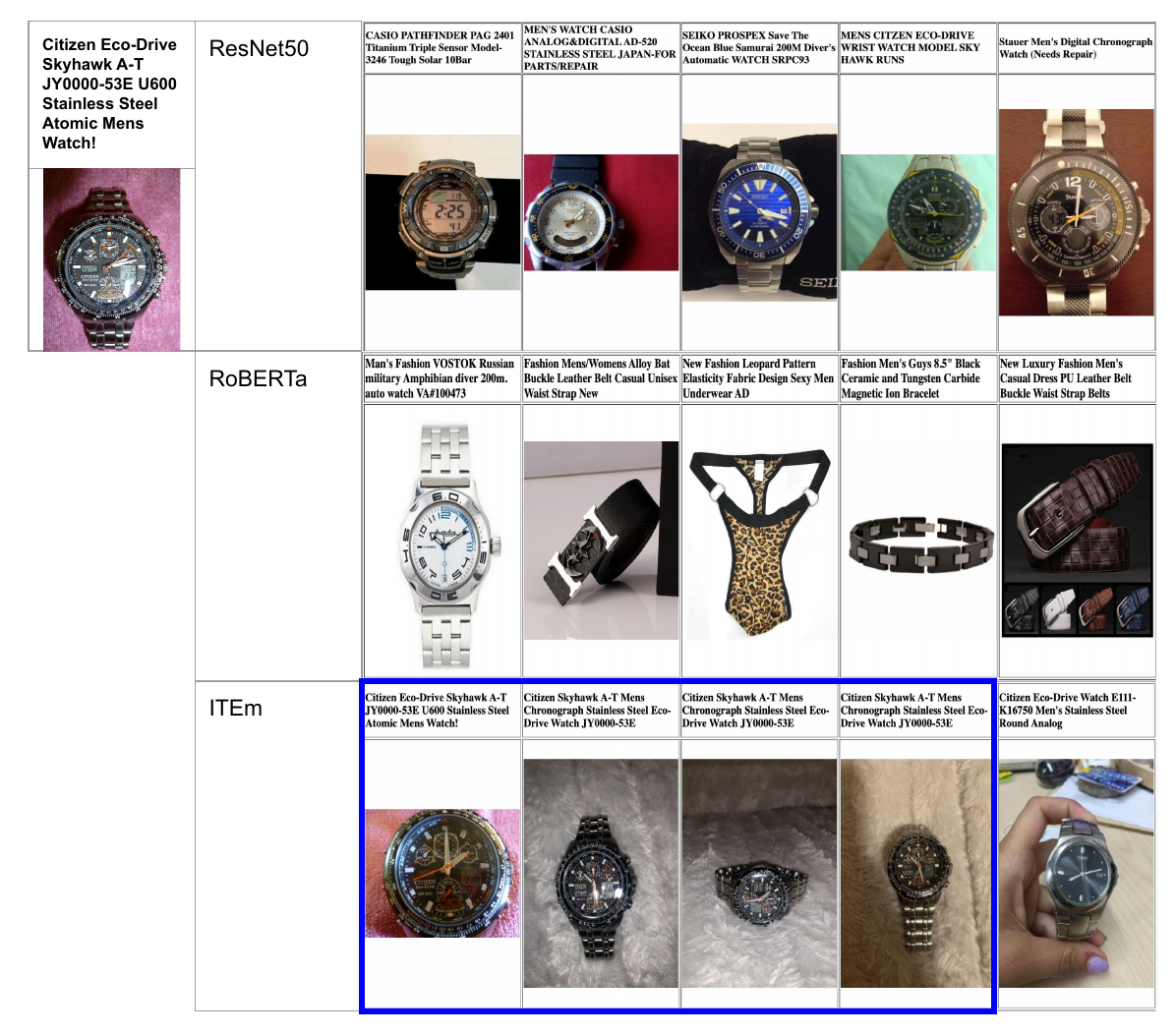}
  \caption{Top 5 examples for the same product recommendation task. The product on the top left is the query. Products in the blue boxes are matching products, while the others are distractors.}
  \label{fig:watch}
\end{figure*}

\begin{figure*}
  \centering
  \includegraphics[width=1\textwidth]{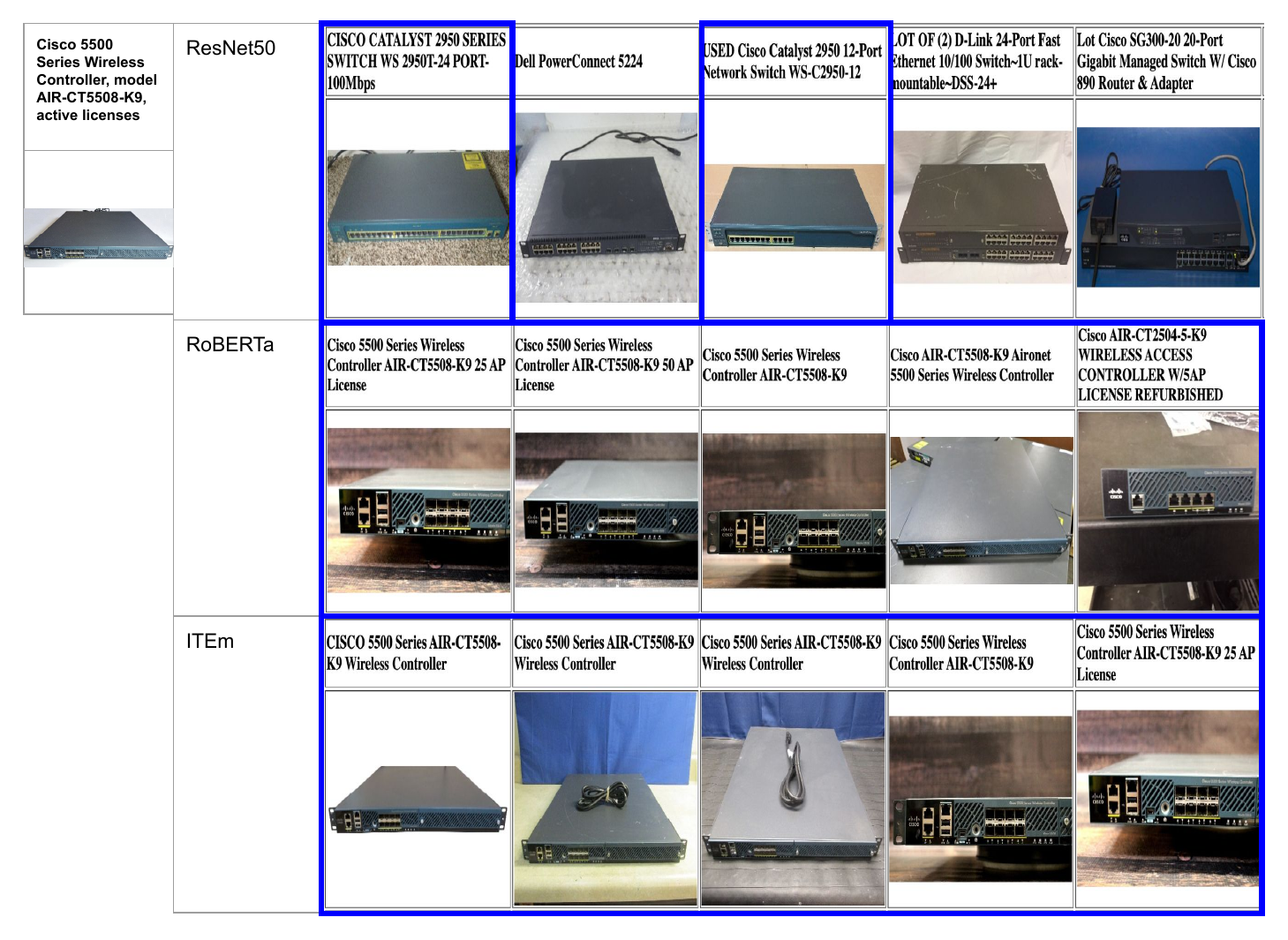}
  \caption{Top 5 examples for the same product recommendation task. The product on the top left is the query. Products in the blue boxes are matching products, while the others are distractors.}
  \label{fig:catalyst}
\end{figure*}

\begin{figure*}
  \centering
  \includegraphics[width=1\textwidth]{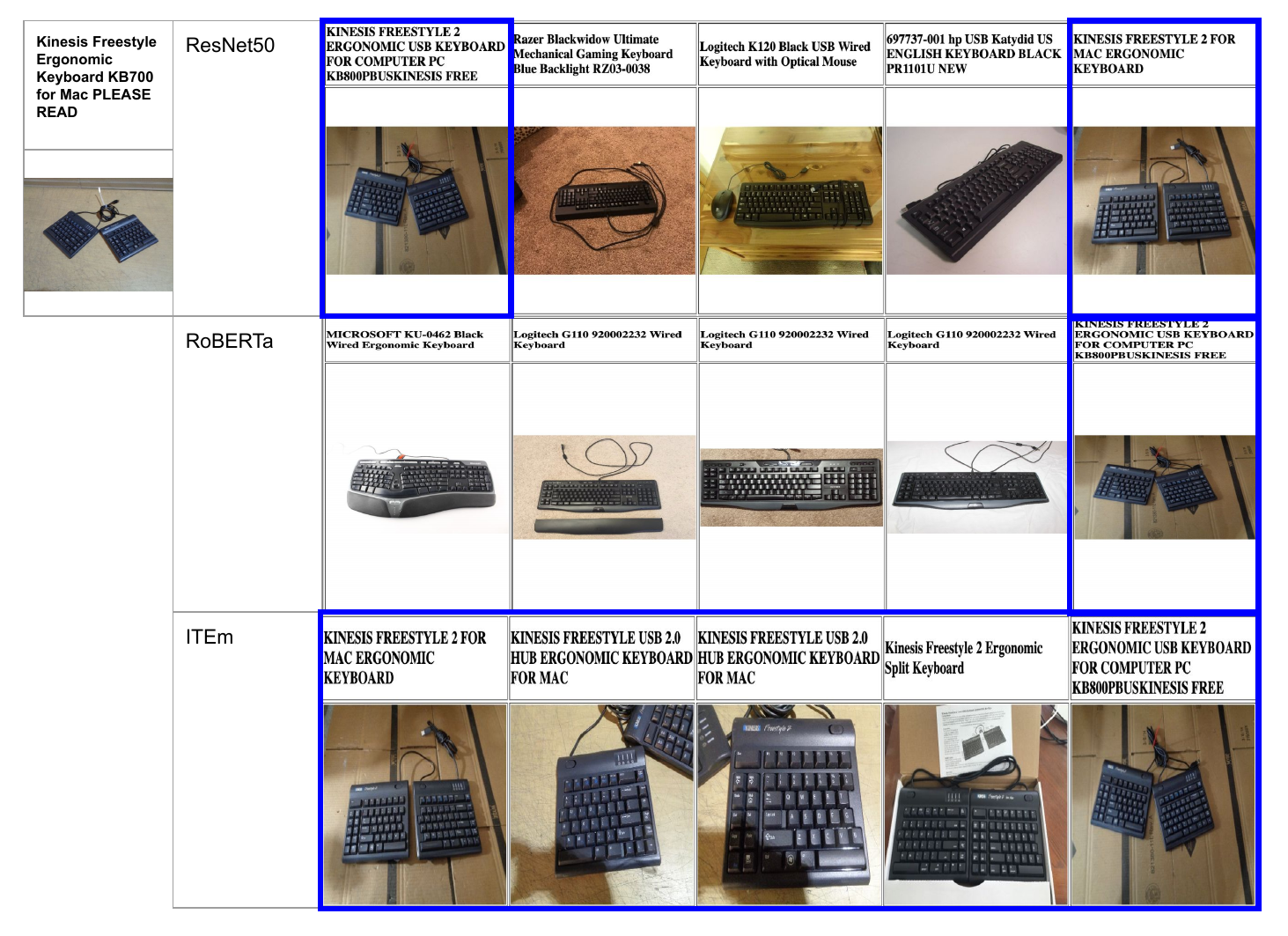}
  \caption{Top 5 examples for the same product recommendation task. The product on the top left is the query. Products in the blue boxes are matching products, while the others are distractors.}
  \label{fig:keyboard}
\end{figure*}

\begin{figure*}
  \centering
  \includegraphics[width=1\textwidth]{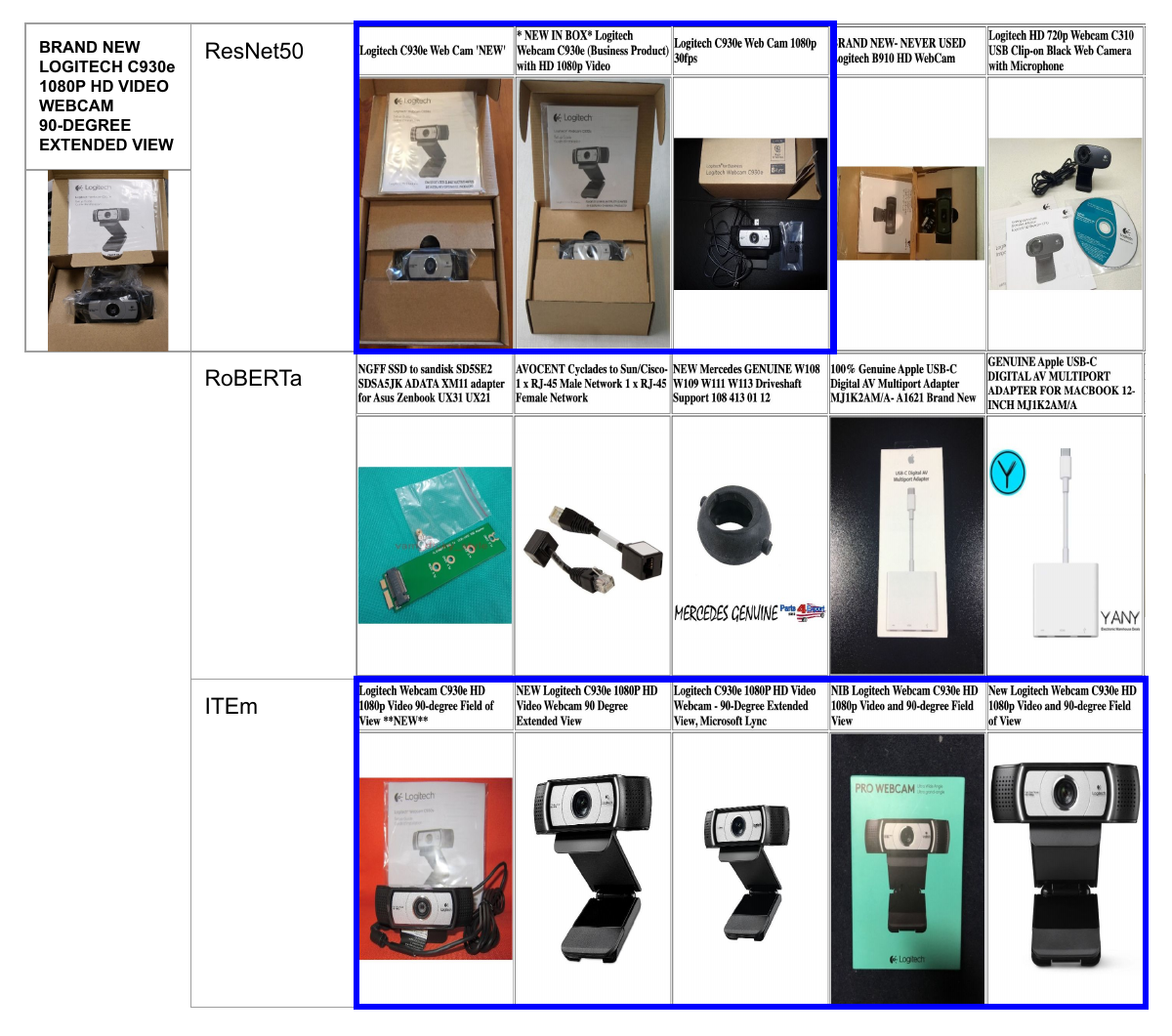}
  \caption{Top 5 examples for the same product recommendation task. The product on the top left is the query. Products in the blue boxes are matching products, while the others are distractors.}
  \label{fig:webcam}
\end{figure*}

Figure \ref{fig:watch}, \ref{fig:catalyst}, \ref{fig:keyboard} and \ref{fig:webcam} show some results of the same product recommendation task. The overall observation is that ITEm tends to provide more diverse matched product than the uni-modal models. For example in Figure \ref{fig:webcam}, ResNet50 puts many attention to the brown paper box. All of the retrieved matched products contain a brown paper box. However, ITEm retrieves products with diverse titles and images. From Figure \ref{fig:keyboard}, we can see that RoBERTa performs well when there are common n-grams in the titles.

\begin{figure*}
  \centering
  \includegraphics[width=1\textwidth]{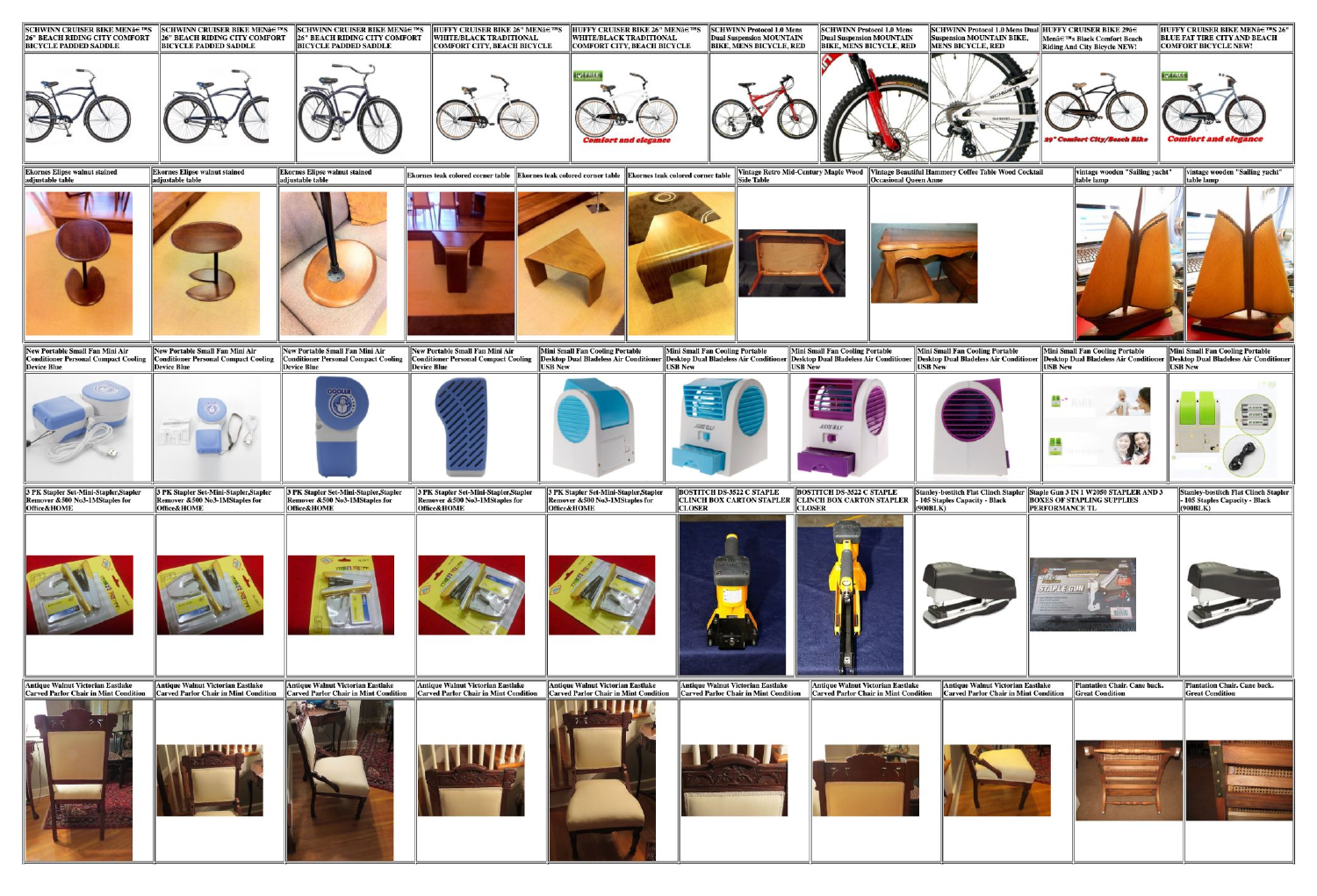}
  \caption{Retrieval result on the SOP test set with the embedding from a pre-trained ITEm. The products in the first column are queries. SOP is not a suitable dataset for the multi-modal retrieval task, because there are too many products with the same titles (caused by the collection strategy of this dataset), making the search rather easy. However, looking at the less similar products, we can still be impressed by the performance of the pre-trained ITEm.}
  \label{fig:sop}
\end{figure*}

There is a public dataset, the Stanford Online Product (SOP) dataset \cite{oh2016deep}, lying in the same domain as our ITOP. We recommend doing some initial experiments on it if you are interested in our method. Figure \ref{fig:sop} shows the retrieval result of our pre-trained ITEm on SOP.

\end{document}